\documentclass[sigconf, nonacm]{acmart}

\AtBeginDocument{%
  }

\usepackage{algorithm}
\usepackage{algorithmic}
\usepackage{booktabs}

\begin{document}

\title{Prompt Sensitivity in Vision-Language Grounding: How Small Changes in Wording Affect Object Detection}

\author{Dawar Jyoti Deka}
\affiliation{%
  \institution{Department of Aerospace Engineering, Indian Institute of Technology Bombay}
  \city{Mumbai}
  \country{India}
}
\email{dawardeka@iitb.ac.in}

\author{Amit Sethi}
\affiliation{%
  \institution{Department of Electrical Engineering, Indian Institute of Technology Bombay}
  \city{Mumbai}
  \country{India}
}
\email{asethi@iitb.ac.in}

\author{Syed Mohammad Ali}
\affiliation{%
  \institution{Department of Electrical Engineering, Indian Institute of Technology Bombay}
  \city{Mumbai}
  \country{India}
}
\email{24b1273@iitb.ac.in}

\renewcommand{\shortauthors}{Deka et al.}

\begin{abstract}
Vision-language models enable open-vocabulary object grounding through natural language queries, under the implicit assumption that semantically equivalent descriptions yield consistent outputs. We examine this assumption using a controlled pipeline combining DETR for object proposals with CLIP for language-conditioned selection on 263 COCO val2017 images. We find that overlapping prompts such as ``a person,'' ``a human,'' and ``a pedestrian'' frequently select different instances, with mean instability of 2.11 distinct selections across six prompts. PCA analysis shows this variability is structured and directional, not random. Prompt ensembling does not improve quality and often shifts selections toward generic regions. We further show that text embedding proximity explains only 34\% of grounding disagreement ($r = -0.58$), confirming that instability arises from the argmax selection mechanism rather than text-level distances alone.
\end{abstract}



\maketitle

\section{Introduction}

Vision-language models such as CLIP~\cite{clip2021} enable visual concepts to be specified via natural language at inference time, bypassing fixed class definitions. Combined with modern detectors like DETR~\cite{detr2020}, they form the backbone of open-vocabulary grounding systems~\cite{glip2022, groundingdino2023, owlvit2022} deployed in robotics, autonomous driving, and multimodal assistants. A central but implicit assumption is that grounding should be stable under semantically equivalent prompts: descriptions such as ``a person,'' ``a human,'' or ``a pedestrian'' are commonly treated as interchangeable~\cite{prompting2021}. If a system's spatial output changes simply because a user says ``human'' instead of ``person,'' this poses a reliability concern for any downstream application.

We conduct an empirical study of prompt sensitivity using a controlled pipeline that decouples object proposal generation from language-conditioned selection. Our experiments show that prompt variation leads to substantial and structured differences in grounding outcomes, and that prompt ensembling~\cite{clip2021}, a common stabilization technique adopted in open-vocabulary detection~\cite{owlvit2022}, does not improve grounding quality and may favor visually generic regions. Our contributions are: (1) empirical documentation of prompt instability across 263 images; (2) geometric analysis revealing structured, directional variability in CLIP similarity space; (3) demonstration that ensembling degrades semantic grounding; and (4) a new analysis showing text embedding proximity explains only 34\% of grounding divergence ($r = -0.58$, $p = 0.024$).

\section{Background and Grounding Pipeline}

DETR~\cite{detr2020} formulates detection as set prediction via a transformer encoder-decoder architecture~\cite{vaswani2017attention}, eliminating hand-designed components such as anchor boxes and non-maximum suppression used in Faster R-CNN~\cite{faster2015} and Mask R-CNN~\cite{mrcnn2017}. It produces deterministic outputs independent of language input. CLIP~\cite{clip2021} aligns images and text in a shared $d$-dimensional embedding space via contrastive learning~\cite{simclr2020, moco2019}, providing two encoders $f_{\text{img}}: \mathcal{I} \to \mathbb{R}^d$ and $f_{\text{text}}: \mathcal{T} \to \mathbb{R}^d$ trained so that paired image-text inputs yield aligned embeddings. In grounding, candidate regions are cropped, encoded by $f_{\text{img}}$ (built on vision transformers~\cite{vit2021} or ResNets trained on ImageNet~\cite{imagenet2009}), and scored against the text embedding $f_{\text{text}}(p)$ via cosine similarity. Earlier vision-language models including ViLBERT~\cite{vilbert2019}, LXMERT~\cite{lxmert2019}, and VisualBERT~\cite{visualbert2019} required task-specific fine-tuning, whereas CLIP enables zero-shot transfer. Open-vocabulary detectors such as MDETR~\cite{mdetr2021}, OWL-ViT~\cite{owlvit2022}, Grounding DINO~\cite{groundingdino2023}, and systems leveraging Open Images~\cite{openimages2020} and VGGNet~\cite{vgg2015} all depend on language-conditioned similarity. Prior work on open-vocabulary detection via captions~\cite{openvocab2021}, robustness to input perturbations~\cite{robustness2019}, explainability~\cite{explainability2016}, and saliency verification~\cite{saliency2018} motivates examining how linguistic variation affects spatial grounding.

Given a set of candidate regions $\mathcal{V}$ produced by DETR and a prompt $p$, grounding reduces to a forced selection problem: exactly one region is chosen per prompt by maximizing cosine similarity. Formally, the selected box for prompt $p$ is
\begin{equation}
\label{eq:sim}
b^{*}(p) = \arg\max_{b \in \mathcal{V}} s(b, p), \quad s(b, p) = \frac{f_{\text{img}}(b)^\top f_{\text{text}}(p)}{\|f_{\text{img}}(b)\| \, \|f_{\text{text}}(p)\|}.
\end{equation}
A decision is always produced regardless of semantic fit, and the output depends entirely on relative rankings rather than absolute correctness. The procedure implementing Equation~\eqref{eq:sim} is formalized in Algorithm~\ref{alg:grounding}. We use DETR ResNet-50 pretrained on COCO; only person-labeled boxes (COCO label index 1) with confidence $\geq 0.5$ are retained, cropped, preprocessed (resize to $224 \times 224$, normalization), encoded by CLIP ViT-B/32, and scored against the prompt embedding. All experiments use identical DETR outputs, isolating the language-conditioned selection stage as the sole variable.

\begin{algorithm}
\caption{CLIP-Conditioned Grounding}
\label{alg:grounding}
\begin{algorithmic}[1]
\REQUIRE Image $I$, prompt $p$, DETR model $D$, CLIP model $C$
\ENSURE Selected bounding box index $b^*$
\STATE $\{(b_i, l_i, c_i)\}_{i=1}^{N} \leftarrow D(I)$
\STATE $\mathcal{V} \leftarrow \{i : l_i = \texttt{person} \wedge c_i \geq 0.5\}$
\STATE $\mathbf{t} \leftarrow f_{\text{text}}(p) / \|f_{\text{text}}(p)\|$
\FOR{each $i \in \mathcal{V}$}
    \STATE $r_i \leftarrow \texttt{crop}(I, b_i)$
    \STATE $\mathbf{v}_i \leftarrow f_{\text{img}}(r_i) / \|f_{\text{img}}(r_i)\|$
    \STATE $s_i \leftarrow \mathbf{v}_i^\top \mathbf{t}$
\ENDFOR
\STATE $b^* \leftarrow \arg\max_{i \in \mathcal{V}} \; s_i$
\RETURN $b^*$
\end{algorithmic}
\end{algorithm}

\section{Experimental Setup}

\textbf{Dataset.} We randomly sample 500 images from COCO val2017~\cite{coco2014} with a fixed seed and retain those with at least one confident person detection, yielding 263 images with an average of 7.0 candidate person boxes per image (range 1--48). Ground-truth annotations are used only for dataset curation.

\textbf{Prompts.} Six prompts target the concept of a human subject: ``a person,'' ``a human,'' ``a woman,'' ``a boy with a hat,'' ``people,'' and ``a pedestrian.'' No prompt tuning is performed.

\textbf{Instability metric.} For each image, we count the number of distinct bounding boxes selected across all six prompts. A value of 1 indicates full agreement; higher values indicate prompt-dependent variation.

\textbf{Ensembling.} Similarity scores are computed independently per prompt and averaged per box; the box with the highest mean score is then selected:
\begin{equation}
\label{eq:ens}
b^{*}_{\text{ens}} = \arg\max_{b \in \mathcal{V}} \frac{1}{K} \sum_{k=1}^{K} s(b, p_k).
\end{equation}
All inference runs on a single NVIDIA L4 GPU in evaluation mode with gradients disabled.

\textbf{Reproducibility.} All experiments use deterministic seeds and pretrained weights from the official DETR and CLIP releases, with PyTorch for inference, NumPy for similarity computation, and scikit-learn for PCA. The pipeline processes 263 images in approximately 12 minutes on a single L4 GPU. No fine-tuning or hyperparameter search is performed, ensuring observed instability reflects intrinsic properties of the pretrained pipeline.

\section{Results}

\subsection{Prompt Instability}

\begin{figure}[t]
\centering
\includegraphics[width=0.95\columnwidth]{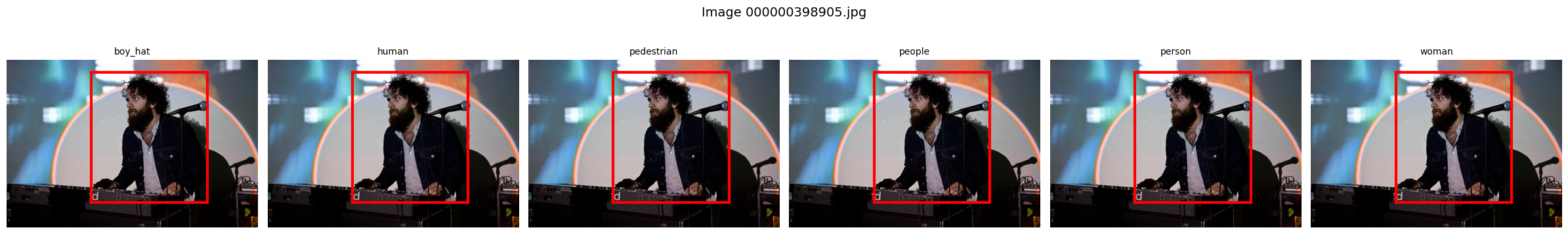}
\caption{Prompt-dependent grounding (example 1). Same DETR proposals grounded with different prompts select different regions despite similar semantics.}
\Description{A row of six images of the same scene, each annotated with a different bounding box selected by a different prompt, showing that the same DETR proposals produce different grounding outputs under different prompts.}
\label{fig:prompt_qual}
\end{figure}

\begin{figure}[t]
\centering
\includegraphics[width=0.95\columnwidth]{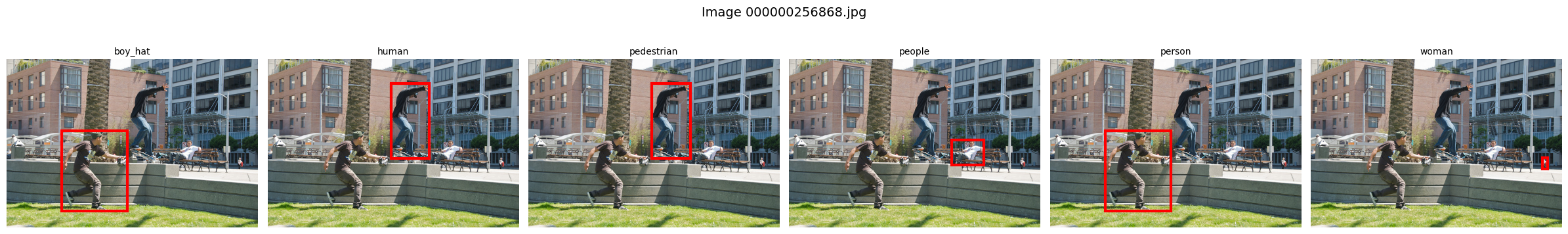}
\caption{Prompt-dependent grounding (example 2). Semantically similar prompts lead to different region selections even in unambiguous scenes with a single dominant subject.}
\Description{A second example showing the same image annotated with different bounding boxes selected by different prompts.}
\label{fig:prompt_qual2}
\end{figure}

Across 263 images, over 55\% exhibit instability greater than one, with mean 2.11 and median 2, meaning semantically overlapping prompts routinely select multiple distinct regions per image. Even ``a person'' and ``a human'' disagree on 17\% of images. These changes are not limited to ambiguous scenes: even with a clearly visible, centrally located person, alternative prompts may select a different individual, a partial body region, or a distant figure. Figures~\ref{fig:prompt_qual} and~\ref{fig:prompt_qual2} show two representative examples.

\begin{figure}[t]
\centering
\includegraphics[width=0.72\columnwidth]{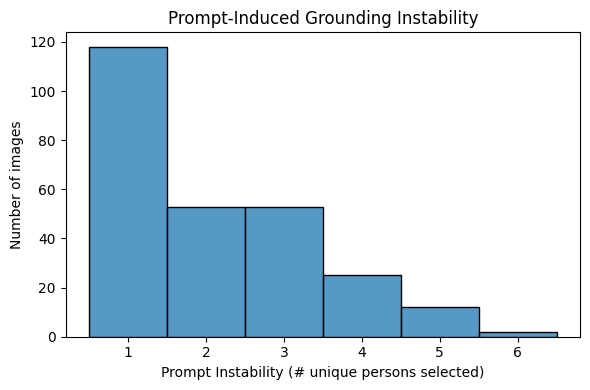}
\caption{Instability distribution over 263 images.}
\Description{A histogram showing the distribution of instability values across 263 images, with most images having 2 distinct selections.}
\label{fig:instability_dist}
\end{figure}

\begin{figure}[t]
\centering
\includegraphics[width=0.9\columnwidth]{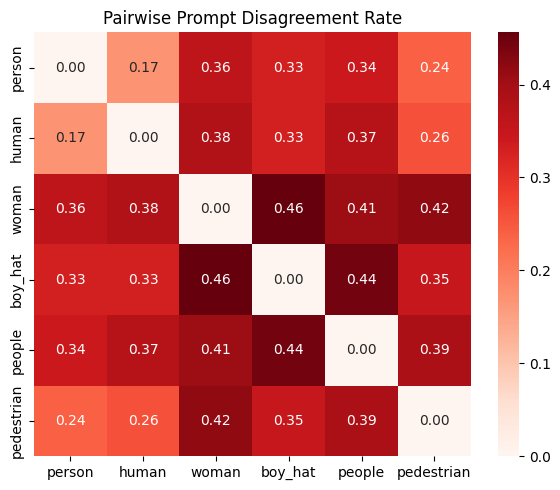}
\caption{Pairwise prompt disagreement rate. Fraction of images where two prompts select different regions.}
\Description{A heatmap showing pairwise disagreement rates between the six prompts.}
\label{fig:pairwise_disagreement}
\end{figure}

The pairwise disagreement matrix (Figure~\ref{fig:pairwise_disagreement}) shows that instability is not uniform: the most stable pair is ``a person'' and ``a human'' (17\% disagreement), while the least stable is ``boy with a hat'' and ``a woman'' (46\%). Intermediate pairs span the range almost uniformly, indicating that disagreement scales smoothly with semantic specificity. Figure~\ref{fig:instability_dist} shows the distribution, with modal value 2 and values of 3 or higher occurring frequently, indicating that for many images at least three of the six prompts select different regions.

\subsection{Geometric Structure in Similarity Space}

\begin{figure}[t]
\centering
\includegraphics[width=0.92\columnwidth]{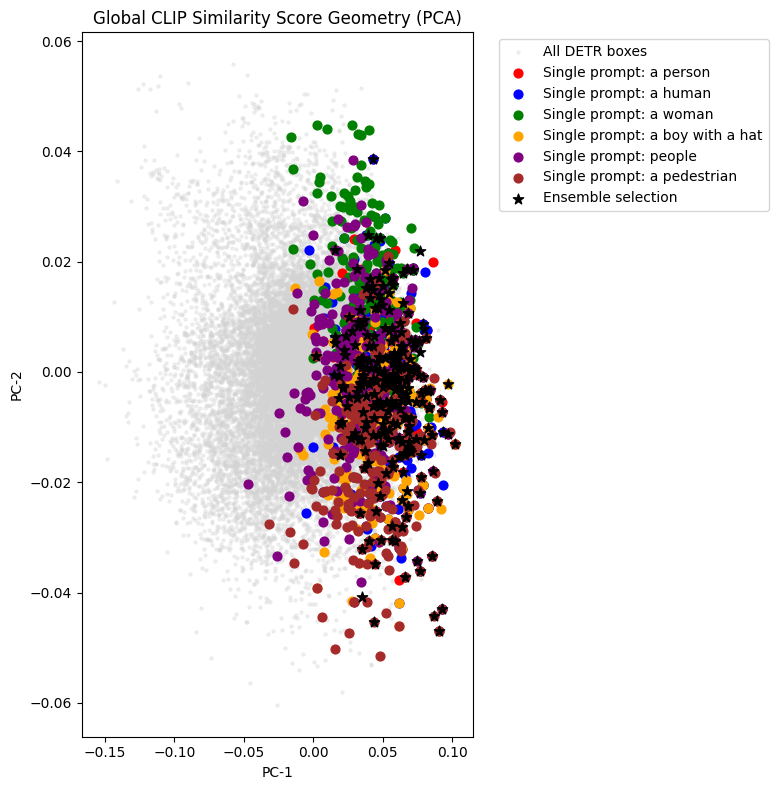}
\caption{PCA projection of CLIP similarity score vectors with prompt-specific selections highlighted. Distinct directional lobes indicate structured variability.}
\Description{A two-dimensional PCA scatter plot showing that selections from different prompts cluster in distinct directional lobes.}
\label{fig:pca_similarity}
\end{figure}

For each image, CLIP similarities between every DETR box and each prompt yield a six-dimensional score vector per box. PCA on these vectors aggregated across all images (Figure~\ref{fig:pca_similarity}) reveals that prompt-specific selections organize into distinct directional lobes rather than scattering isotropically. The first two principal components capture the dominant axes along which similarity responses vary, and selections from different prompts consistently occupy different angular regions of this space. If variability arose from noise, selections would scatter without orientation; instead, prompts emphasize different semantic cues, producing coherent geometric patterns that persist across images. Notably, the lobe associated with ``a boy with a hat'' is angularly farther from the others, reflecting its more specific visual constraints, while ``a person'' and ``a human'' produce overlapping but non-identical lobes, consistent with their high text similarity but non-zero grounding disagreement.

\begin{figure}[t]
\centering
\includegraphics[width=0.78\columnwidth]{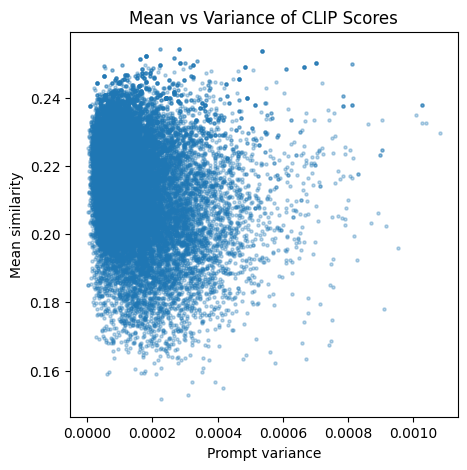}
\caption{Mean vs.\ variance of CLIP similarity scores. High similarity does not imply prompt stability.}
\Description{A scatter plot of mean similarity versus prompt-induced variance for all candidate regions, showing that high-similarity regions can have high variance.}
\label{fig:mean_variance}
\end{figure}

Plotting per-box mean similarity against prompt-induced variance (Figure~\ref{fig:mean_variance}) reveals a decoupling between confidence and stability: many high-mean regions also exhibit high variance, indicating strong alignment with some prompts but weak alignment with others. Conversely, some regions show low variance despite modest similarity scores, reflecting consistent but less distinctive alignment. Confidence in a grounding output does not imply robustness to prompt reformulation, and selecting boxes purely by maximum similarity provides no guarantee of stability across linguistically equivalent queries.

\subsection{Why Prompt Ensembling Fails}

Prompt ensembling was introduced by Radford et al.~\cite{clip2021} for zero-shot classification and adopted in open-vocabulary detection~\cite{owlvit2022, glip2022}. Defined in Equation~\eqref{eq:ens}, it implicitly assumes that variation across prompts reflects unstructured noise that can be reduced by averaging. Our analysis shows the opposite. For many regions, the prompt-induced variance
\begin{equation}
\sigma^2(b) = \frac{1}{K} \sum_{k=1}^{K} \left( s(b, p_k) - \bar{s}(b) \right)^2
\end{equation}
is high precisely because the region aligns strongly with some prompts and weakly with others, where $\bar{s}(b) = \frac{1}{K}\sum_k s(b, p_k)$. Averaging suppresses $\sigma^2(b)$ by construction, systematically penalizing semantically specific regions in favor of those with moderate but uniform similarity. The ensemble selection thus becomes biased toward visually generic regions that match all prompts equally well.

\begin{figure}[t]
\centering
\includegraphics[width=0.46\columnwidth]{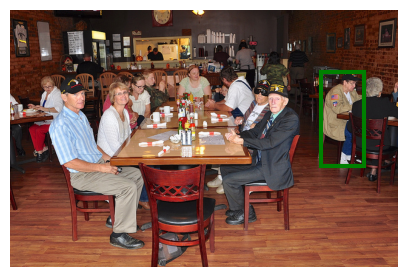}
\includegraphics[width=0.46\columnwidth]{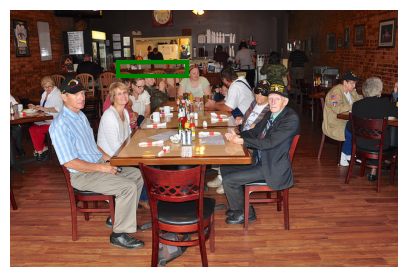}
\caption{Ensembling failure (example 1). Left: single prompt correctly grounds to person. Right: ensemble shifts selection toward a visually generic region.}
\Description{Two side-by-side images showing a correct single-prompt grounding on the left and an incorrect ensemble grounding on the right.}
\label{fig:ensemble_failure}
\end{figure}

\begin{figure}[t]
\centering
\includegraphics[width=0.46\columnwidth]{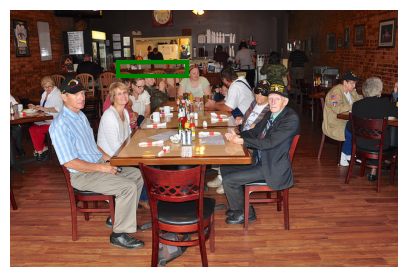}
\includegraphics[width=0.46\columnwidth]{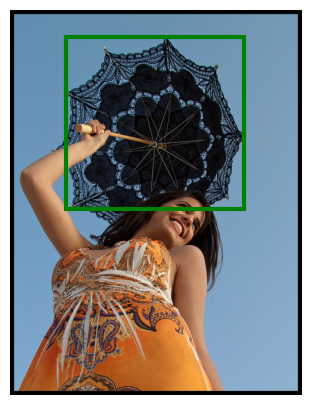}
\caption{Ensembling failure (example 2). Ensemble averaging degrades grounding quality by favoring prompt-agnostic regions over semantically specific ones.}
\Description{A second comparison showing how ensemble averaging degrades grounding quality.}
\label{fig:ensemble_failure2}
\end{figure}

In practice, ensembling shifts grounding toward partial objects or background regions when individual prompts correctly ground to human instances (Figures~\ref{fig:ensemble_failure} and~\ref{fig:ensemble_failure2}). Stability alone is an inadequate objective if it is achieved by suppressing semantic structure: a robust grounding mechanism must distinguish noise-variance from signal-variance, treating only the former as undesirable. Designing such mechanisms requires moving beyond uniform averaging toward weighted or learned aggregation schemes that preserve prompt-specific information.

\subsection{Text Embedding Proximity vs.\ Grounding Divergence}

\begin{figure}[t]
\centering
\includegraphics[width=0.88\columnwidth]{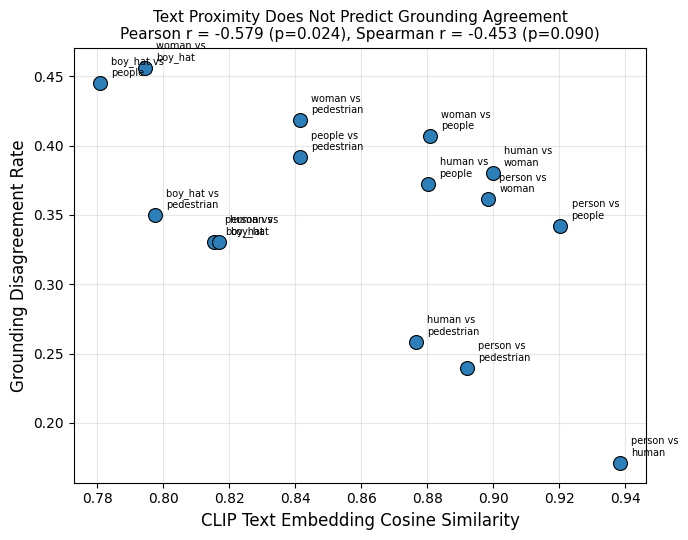}
\caption{Text embedding cosine similarity vs.\ grounding disagreement for 15 prompt pairs. The moderate correlation ($r = -0.58$) confirms that text proximity alone cannot predict grounding stability.}
\Description{A scatter plot of text embedding similarity versus grounding disagreement for all 15 prompt pairs.}
\label{fig:scatter_textsim}
\end{figure}

A natural question is whether instability simply reflects distances between prompt embeddings in CLIP's text space. We compute pairwise cosine similarity between the CLIP text embeddings of all six prompts and compare against grounding disagreement (Table~\ref{tab:text_vs_ground}, Figure~\ref{fig:scatter_textsim}).

\begin{table}[t]
\centering
\caption{CLIP text embedding cosine similarity (upper triangle) vs.\ pairwise grounding disagreement rate (lower triangle). Pearson $r = -0.58$, $p = 0.024$.}
\label{tab:text_vs_ground}
\small
\setlength{\tabcolsep}{3pt}
\begin{tabular}{lcccccc}
\toprule
 & pers. & hum. & wom. & boy+hat & ppl. & ped. \\
\midrule
pers. & -- & \textbf{0.94} & \textbf{0.90} & \textbf{0.82} & \textbf{0.92} & \textbf{0.89} \\
hum. & 0.17 & -- & \textbf{0.90} & \textbf{0.82} & \textbf{0.88} & \textbf{0.88} \\
wom. & 0.36 & 0.38 & -- & \textbf{0.79} & \textbf{0.88} & \textbf{0.84} \\
boy+hat & 0.33 & 0.33 & 0.46 & -- & \textbf{0.78} & \textbf{0.80} \\
ppl. & 0.34 & 0.37 & 0.41 & 0.44 & -- & \textbf{0.84} \\
ped. & 0.24 & 0.26 & 0.42 & 0.35 & 0.39 & -- \\
\bottomrule
\end{tabular}
\end{table}

The Pearson correlation between text similarity and grounding disagreement is $r = -0.58$ with $p = 0.024$, explaining only 34\% of the variance ($r^2 = 0.34$). Several pairs illustrate this gap: ``a person'' and ``a human'' share text similarity 0.94 yet disagree on 17\% of images, while ``people'' and ``a pedestrian'' have lower similarity (0.84) but disagree at 39\%. Because grounding operates via argmax over candidate regions, small perturbations in the similarity score vector $\{s(b, p)\}_{b \in \mathcal{V}}$ can produce discrete jumps in the selected box whenever the top-ranked candidate changes. This non-Lipschitz behavior means that bounded changes in the input prompt embedding can produce arbitrarily large changes in the grounded region, even when the text embeddings themselves remain close. Text embedding proximity alone is therefore insufficient to predict grounding stability; the directional patterns revealed by PCA reflect prompt-specific visual biases that emerge at the selection stage, not the text encoding stage.

\section{Discussion}

\subsection{Summary of Findings}

Our experiments reveal three principal findings. First, prompt invariance does not hold in similarity-based grounding: over 55\% of images produce different grounding outcomes under semantically overlapping prompts, with a mean of 2.11 distinct selections per image. Second, this variability is structured rather than random, with different prompts consistently biasing selections toward distinct visual cues, as revealed by directional lobes in the PCA projection of similarity score vectors. Third, prompt ensembling, the most commonly adopted mitigation strategy, can degrade grounding quality by suppressing semantically meaningful variance, and text embedding proximity accounts for only 34\% of the observed divergence ($r = -0.58$, $p = 0.024$).

\subsection{Practical Implications}

These findings have practical consequences for deployed systems. In robotics and autonomous perception, prompt sensitivity means that the choice of natural language instruction can alter which physical object a system attends to, affecting downstream manipulation or navigation. A robot instructed to ``pick up the person's cup'' versus ``pick up the human's cup'' may attend to different individuals in the scene, leading to incorrect actions. In multimodal retrieval and visual question answering, equivalent query reformulations may produce inconsistent results, undermining user trust. In safety-critical applications such as autonomous driving, where pedestrian detection may be conditioned on language, the system could respond differently to ``a pedestrian crossing the street'' versus ``a person walking across the road,'' with potentially serious consequences. The issue extends to medical imaging, where descriptions of anatomical regions may vary across clinicians, and to content moderation, where similar policy descriptions may yield inconsistent flagging.

The argmax selection mechanism is the root cause: it amplifies small perturbations in similarity scores into discrete changes in spatial output, creating a brittleness invisible at the text embedding level but consequential at the grounding level. Systems relying on CLIP-based grounding should evaluate performance across multiple prompt formulations and report variance, not just best-case results, since reporting only the best-case prompt overstates real-world performance.

\subsection{Limitations and Future Work}

Our study evaluates a single pipeline (DETR ResNet-50 with CLIP ViT-B/32); whether end-to-end open-vocabulary detectors such as Grounding DINO~\cite{groundingdino2023} or OWL-ViT~\cite{owlvit2022} exhibit similar sensitivity remains open, since these systems incorporate language conditioning during detection rather than as a post-hoc selection step. Future work should separate near-synonyms from semantically distinct prompts, incorporate IoU-based evaluation against ground-truth annotations, and investigate whether larger CLIP backbones (ViT-L/14, ViT-B/16) exhibit reduced brittleness. More broadly, grounding systems require structural constraints beyond similarity aggregation.

\section{Conclusion}

We studied how small changes in prompt wording affect vision-language grounding. Using a controlled two-stage pipeline with DETR and CLIP on 263 COCO images, we found that semantically similar prompts frequently select different objects, with instability occurring in over half of all images and a mean of 2.11 distinct selections per image across six prompts. This variation is not random noise but follows consistent, structured patterns where different prompts emphasize different visual features, as confirmed by the directional lobes observed in the PCA projection of similarity score vectors. Prompt ensembling, the standard mitigation strategy, reduces variation but does not improve accuracy and often selects semantically worse regions by suppressing meaningful prompt-specific signal alongside genuine noise.

Our analysis of text embedding proximity reveals that only 34\% of grounding disagreement can be attributed to text-level distance ($r = -0.58$, $p = 0.024$), confirming that the argmax selection mechanism is the primary source of brittleness rather than the text encoder itself. These observations highlight fundamental limitations of current similarity-based grounding approaches and motivate the development of grounding systems with structural constraints that preserve semantic specificity while ensuring prompt-robust behavior.

\begin{acks}
The authors thank the reviewers for their valuable feedback and suggestions that helped improve this paper.
\end{acks}

\end{document}